\documentclass[10pt,twocolumn,letterpaper]{article}

\usepackage[pagenumbers]{cvpr} 
\usepackage{stfloats}
\usepackage[dvipsnames]{xcolor}

\definecolor{cvprblue}{rgb}{0.21,0.49,0.74}
\usepackage[pagebackref,breaklinks,colorlinks,citecolor=cvprblue]{hyperref}

\definecolor{GQL}{RGB}{45,149,124}

\def\paperID{13649} 
\def\confName{CVPR}
\def\confYear{2024}

\title{IA2U: A Transfer Plugin with Multi-Prior for In-Air Model to Underwater}

\author{Jingchun Zhou\textsuperscript{\rm 1},
Qilin Gai\textsuperscript{\rm 1}, 
Kin-man Lam\textsuperscript{\rm 2}, 
Xianping Fu\textsuperscript{\rm 1}
\\
\textsuperscript{\rm 1} Dalian Maritime University, 
\textsuperscript{\rm 2} Hong Kong Polytechnic University
}

\begin{document}
\maketitle
\begin{abstract}
In underwater environments, variations in suspended particle concentration and turbidity cause severe image degradation, posing significant challenges to image enhancement (IE) and object detection (OD) tasks. Currently, in-air image enhancement and detection methods have made notable progress, but their application in underwater conditions is limited due to the complexity and variability of these environments. Fine-tuning in-air models saves high overhead and has more optional reference work than building an underwater model from scratch. 
To address these issues, we design a transfer plugin with multiple priors for converting in-air models to underwater applications, named IA2U. IA2U enables efficient application in underwater scenarios, thereby improving performance in Underwater IE and OD. IA2U integrates three types of underwater priors: the water type prior that characterizes the degree of image degradation, such as color and visibility; the degradation prior, focusing on differences in details and textures; and the sample prior, considering the environmental conditions at the time of capture and the characteristics of the photographed object. Utilizing a Transformer-like structure, IA2U employs these priors as query conditions and a joint task loss function to achieve hierarchical enhancement of task-level underwater image features, therefore considering the requirements of two different tasks, IE and OD. Experimental results show that IA2U combined with an in-air model can achieve superior performance in underwater image enhancement and object detection tasks. The code will be made publicly available.
\end{abstract}   
\section{Introduction}
\label{sec:intro}

\begin{figure}
    \centering
    \includegraphics[width=0.8\linewidth, height=0.9\linewidth]{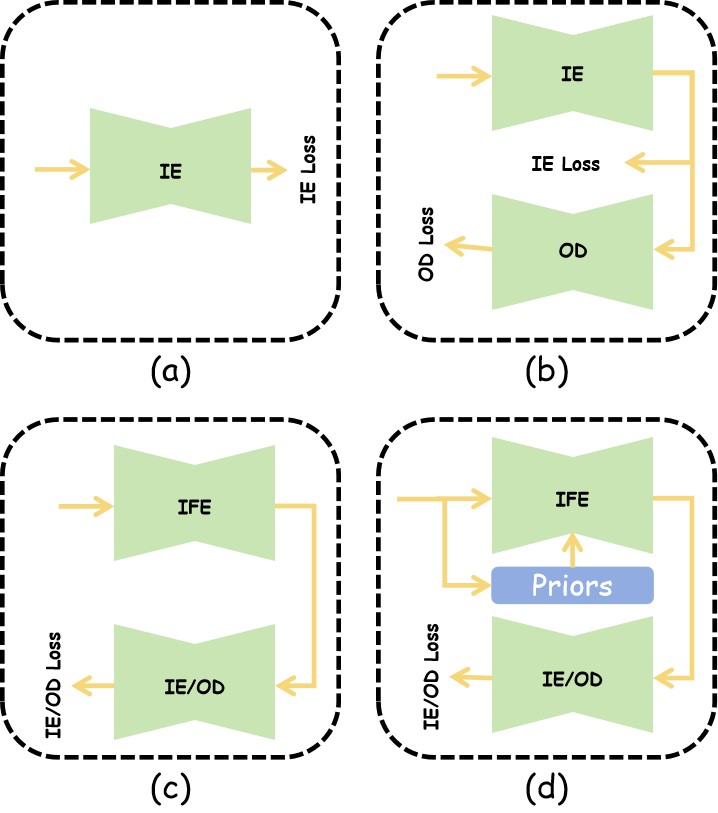}
    \caption{Illustrate the overall process of (a) image enhancement, (b) task-driven image enhancement, (c) feature enhancement, and (d) proposed IA2U. In particular, the \textbf{IE} for image enhancement. \textbf{IFE} presents image feature enhancement. \textbf{OD} for object detection, and \textbf{IE/OD Loss} is the loss function for the corresponding task.}
    \label{fig:diff}
\end{figure}

Underwater image quality degradation due to light attenuation in water is a significant barrier to the development of underwater vision. The presence of color shifts and low contrast in underwater images not only reduces visual perception for the human eye but also increases the difficulty of tasks such as object detection. As shown in Fig.\ref{fig:diff} (a), UIE methods~\cite{LI2020107038, 9930878, zhou2023underwater, 10177702, huang2023contrastive,URanker} are used to improve degraded visual quality. In recent years, many advanced UIE methods have been proposed, utilizing neural networks to enhance low-quality underwater images to be closer to clear reference images. Compared to UIE, there is relatively less research in Underwater Object Detection (UOD). Currently, the UOD methods can be categorized into three types: \textbf{(1). Enhance-then-Detect Method:} UIE is used to improve the visual quality of underwater images, followed by OD using the enhanced images. The UIE and UOD models are trained independently. However, ~\cite{hashmi2023featenhancer} has shown that image enhancement may negatively impact OD; \textbf{(2). In-Air Model Transfer Method:} Applying terrestrial OD models directly to underwater environments is limited due to the lack of specific underwater environmental information; \textbf{(3). Integrated Task-Driven Method:} As shown in Fig. \ref{fig:diff} (b), unlike traditional cascading methods, the task-driven UIE~\cite{9832540} trains UIE and UOD models simultaneously. This integrated approach aims to balance visual quality and OD performance, but joint training with multi-task loss increases the difficulty of model convergence. The above methods present two key challenges: \textbf{(1)}. How do we adapt in-air models to the underwater environment using fine-tuning to reduce the high overhead of designing an underwater model from scratch?
\textbf{(2)}. How to simultaneously optimize image visual quality and object detection performance?

To address these challenges, we are inspired by FeatEnhancer~\cite{hashmi2023featenhancer}, which proposes an Image Feature Enhancement (IFE) method, as illustrated in Fig.\ref{fig:diff} (c). The IFE approach adaptively enhances task-level features by combining them with hierarchical features guided by task loss, thereby improving the performance of IE and OD. Regardless, underwater images' degradation representation differs significantly from in-air images. Classical IFE methods cannot identify this representation difference while accommodating upstream and downstream tasks, leading to impaired performance when applied to underwater scenes. To overcome the above problems,  we design a novel plug-and-play underwater plugin called IA2U, as shown in Fig.\ref{fig:diff} (d). IA2U utilizes RGB images as input and employs a pre-trained water-type classifier to predict the characteristics of the input images. The predictions of the classifier, along with intermediate layer features, provide a wealth of a priori knowledge. We utilize this knowledge for deep feature retrieval, capturing extensive information about the underwater environment. The retrieved features are refined using a full-scale feature alignment strategy and hierarchically aggregated with a Transformer-like structure, enhancing task-relevant features under the guidance of IE or OD loss functions. Unlike existing methods, IA2U is a plug-and-play image feature enhancement plugin positioned before the IE or OD model, embedding more underwater environmental information into the in-air model for improved performance in underwater datasets. Additionally, IA2U can handle both IE and OD tasks by modifying the types of in-air models and loss functions. Our main contributions are summarized as follows:

(1) We propose an innovative full-scale feature aggregation module that abandons the traditional single-anchor alignment approach, effectively reducing information loss during the feature fusion process. This optimizes the efficiency and accuracy of feature fusion.

(2) We design IA2U, a plug-and-play IFE plugin specifically for underwater environments. IA2U integrates various underwater priors, encompassing extensive features of the underwater environment, and effectively adapts in-air models to underwater scenarios, significantly improving performance in IE and OD tasks.

(3) Extensive experiments demonstrate that IA2U significantly enhances the performance of in-air models for UIE and OD. Its plug-and-play nature allows easy adaptation to various models, showcasing its broad applicability and flexibility in diverse applications.

\section{Related Work}

\subsection{Underwater Image Enhancement}
UIE methods aim to improve underwater image quality by utilizing underwater imaging models, adjusting pixel values, or learning underwater features. These methods are primarily categorized into three types: model-based methods, model-free methods, and data-driven methods. Model-based methods~\cite{9854113, 9257110, 9788535} typically employ simplified underwater imaging models~\cite{akkaynak2019sea, 50695} and use prior knowledge to estimate the transmission map and background light, resulting in clearer images. However, in complex underwater environments, reliance on single prior knowledge leads to erroneous parameter estimation and accumulated errors, affecting image restoration quality. Model-free methods~\cite{TEBCF, 9895452} process image color and brightness information using pixel statistics, adjusting pixel values to fit a specific distribution. Yet, these methods apply a uniform treatment to all pixels, leading to over- or under-enhancement in certain image areas. Data-driven methods, such as Ucolor~\cite{9426457}, combine underwater optical imaging models with deep learning technology, embedding multi-color space tokens to acquire more degradation information. UGIF-Net~\cite{10177702} extensively extracts image color information through a dense attention mechanism and addresses color drift issues through feature fusion. U-shape~\cite{10129222} introduces Transformer~\cite{vaswani2017attention} structures into UIE, recovering detailed information in images through global modeling capabilities. However, due to the complexity of underwater environments and the difficulty in obtaining paired data, data-driven methods have limited adaptability in multi-water scenarios with mixed degradation characteristics. Therefore, there need to design robust water degradation priors to enhance the model's adaptability in various underwater environments.

\label{sec:related_work}
\begin{figure*}[t]
    \centering
    \includegraphics[width=0.9\linewidth]{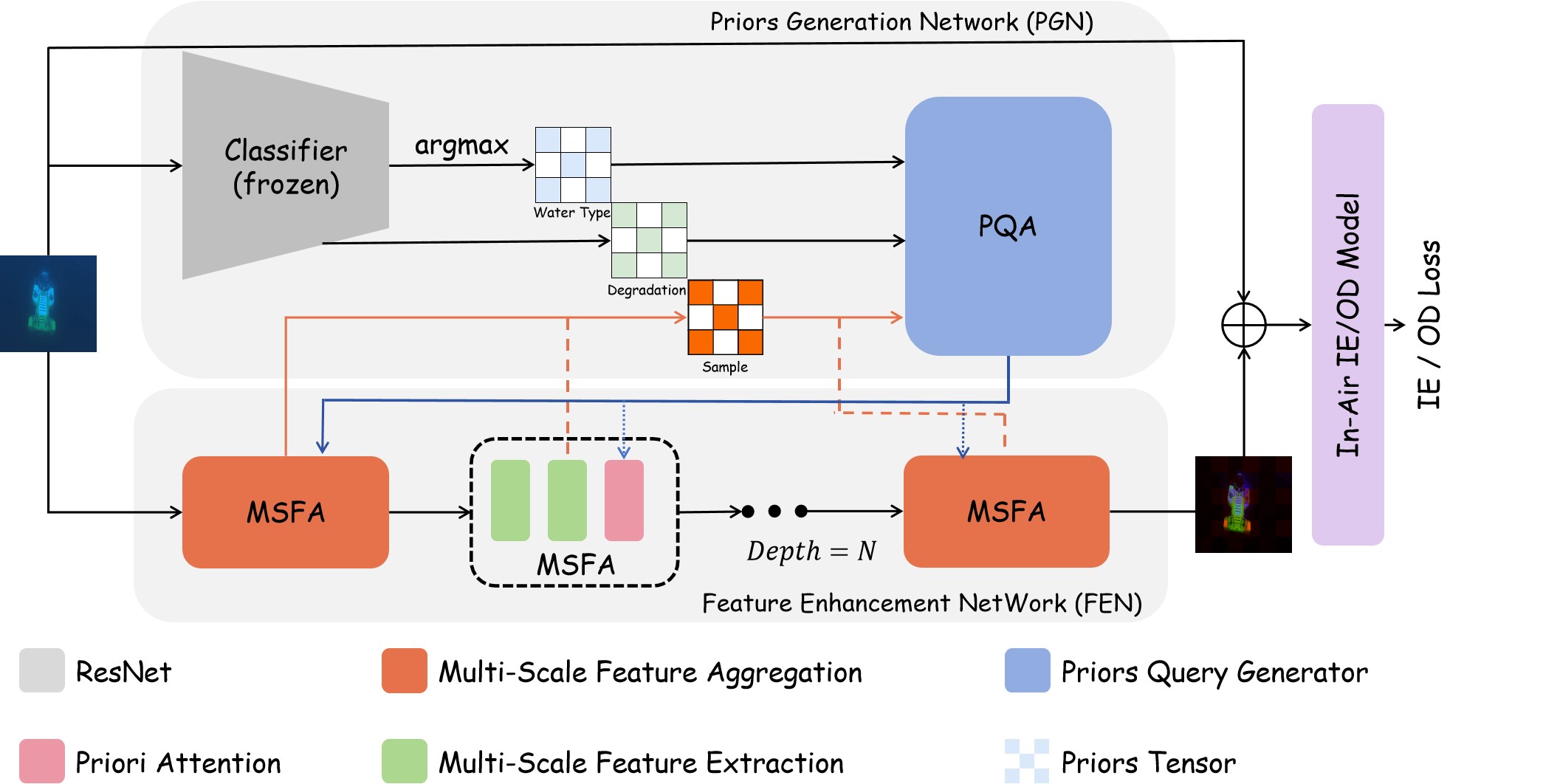}
    \caption{Overview architecture of the IA2U method. The underwater image is passed through the classifier to obtain two prior knowledge of water type and degradation features and combined with the features in FEN to generate Query. Next, the generated Query is fed to the Attention module in FEN for feature refinement. Finally, the feature increments are combined with the original image through the residual connection to feed the subsequent IE and OD network.}
    \label{fig:model}
\end{figure*}

\subsection{Object Detection}
In the field of UOD, Faster RCNN~\cite{7485869} marks a milestone by being the first to integrate feature extraction, bounding box regression, and classification into a single model, optimizing generation speed through region proposal networks and achieving a balance between performance and efficiency. Subsequently, RetinaNet~\cite{8237586} addressed the issue of category imbalance with focal loss, ushering single-stage OD networks into a new phase of development. The advent of FCOS~\cite{9010746} further simplified the OD process by eliminating anchors, thus avoiding complex boundary matching and effectively enhancing performance in other visual tasks. Moreover, TOOD~\cite{9710724} strengthened the interaction between classification and localization tasks through its task alignment head, leading to more accurate detection outcomes. Despite these advancements, these methods still face challenges in UOD. Compared to in-air OD, UOD suffers due to the complexity of the environment and scarcity of data, leading to limited performance. Currently, most UIE methods are primarily optimized for human-eye perception, which proves insufficiently effective in UOD tasks, sometimes even resulting in adverse effects. Therefore, adapting and transferring these efficient in-air models to underwater environments to enhance their performance in underwater datasets becomes a crucial problem that urgently needs to be addressed.

\subsection{Image Feature Enhancement}
IFE is focused on amplifying task-specific features at the task level. Unlike traditional IE methods, IFE exhibits greater flexibility by enhancing different features based on various task losses, instead of relying solely on image reconstruction loss for model convergence. This approach allows IFE to adapt and respond more effectively to the specific requirements of different tasks. For example, TACL~\cite{9832540} follows a similar IFE approach by combining IE and OD losses to constrain the model, aiming to balance visual quality and detection performance. FeatEnhancer~\cite{hashmi2023featenhancer} represents a more advanced IFE method that significantly enhances the performance of both upstream and downstream tasks through the strengthening of hierarchical features. Nevertheless, these methods still face numerous challenges when applied to complex and diverse underwater environments, especially in adapting to varied underwater scenes while satisfying both IE and OD tasks. Therefore, designing an effective plugin or module that renders in-air models more robust and efficient in underwater environments has become an urgent issue to address.

\section{Proposed method}
The IA2U plugin primarily addresses the challenge of effectively applying in-air vision models to underwater scenes, while simultaneously supporting UIE and UOD tasks. As shown in Figure \ref{fig:model}, IA2U adopts a dual-branch architecture, comprising a Feature Enhancement Network (FEN) and an Underwater Prior Generation Network (PGN). The PGN integrates a pre-trained classifier, and the input image is fed in parallel to both branch networks. In the FEN, a multi-scale feature extraction module is used to refine the image features. These refined features are passed to the PGN as the sample prior. In the PGN, the input image is processed through a classifier to determine the water type and related degradation characteristics. Based on these priors and sample features, the PGN generates a `Query', which is then fed back to the attention module of the FEN, achieving effective integration of features and prior information. Through multiple iterations, the FEN combines the enhanced features with the original input image using a residual structure, generating the final enhanced image. This process provides more precise and robust feature information for IE and OD.

\begin{figure}[h]
    \centering
    \includegraphics[width=0.8\linewidth]{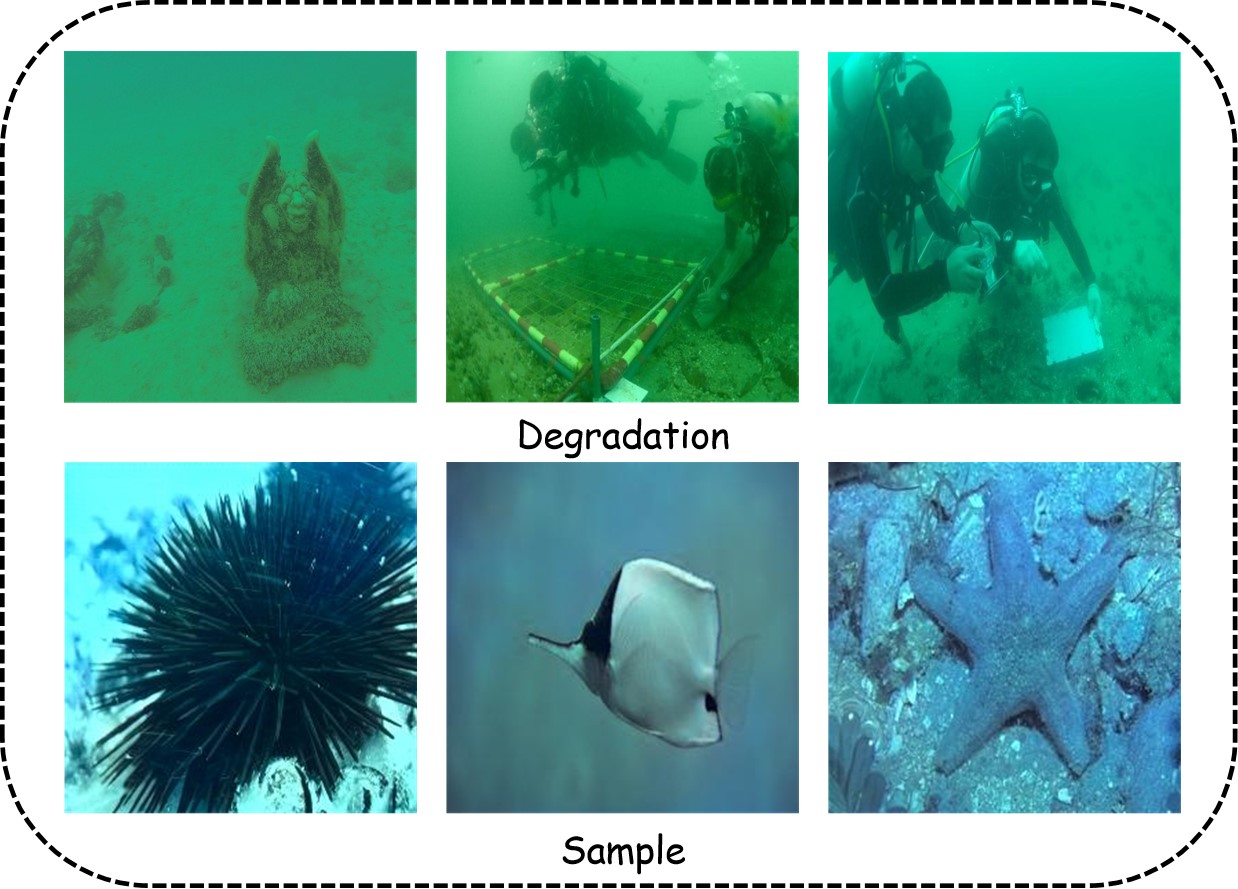}
    \caption{Prior Selection in IA2U Model. Top row: Degradation prior knowledge in various water types, emphasizing different degradation levels. Bottom row: Sample prior knowledge, categorized by differences in ambient lighting and object characteristics.}
    \label{fig:priori}
\end{figure}

\subsection{Multiple underwater priori knowledge}
{\bf Priori knowledge design:} 
In underwater vision tasks, setting appropriate the prior knowledge is one of the key challenges. ~\cite{zhou2023underwater, TEBCF, 9854113} has shown that too strong the prior constraints can lead to model failure. Therefore, the  prior knowledge should not only encompass underwater environmental information but also balance the constraining power. This study adopts three types of weaker priors, as follows: {\bf (1). Water type prior knowledge:} The degradation of underwater images is primarily caused by the attenuation of light in water, and different water bodies have varying rates of light attenuation ~\cite{berman2017diving}. Therefore, the degree of image degradation is directly related to the water type. {\bf (2). Degradation prior knowledge:} Even within the same type of water body, there are differences in the degree of image degradation due to minor differences in suspended particle concentration and other factors. This variation is exemplified in the top row of seabed images in \cref{fig:priori}. {\bf (3). Sample prior knowledge:} As shown in the bottom of \cref{fig:priori}, factors such as the intensity of light, the radiation rate of the lighting object, and the shooting location also affect the image, exhibiting unique characteristic performances. This is referred to as sample prior knowledge.

\begin{figure}[htbp]
    \centering
    \includegraphics[width=0.7\linewidth, scale=0.4]{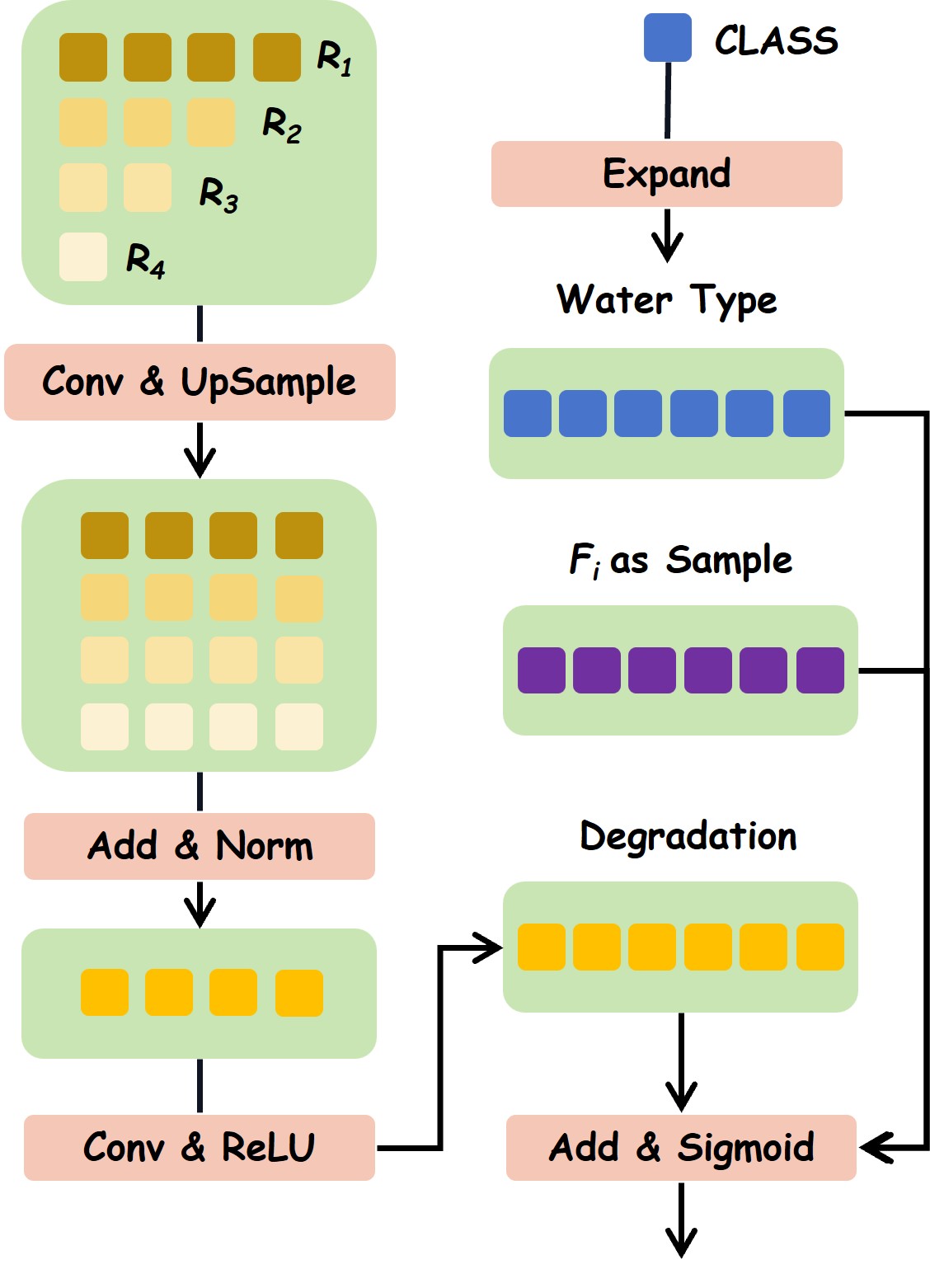}
    \caption{Structure of the Priori Query Generator (PQG). Depicting classifier predictions (CLASS) and middle layer features (R), with $F_{j}$ representing the output of the $j$th layer of the FEN network}
    \label{fig:priori_gen}
\end{figure}

{\bf Generation of multiple prior knowledge:} \cref{fig:priori_gen} illustrates the process for generating water-type prior knowledge from a given low-quality underwater image $X$. This is achieved by Eq. (\ref{eq:1}):
\begin{equation}
    Prob = R(X, \theta_{frozen})
        \label{eq:1}
\end{equation}
where $Prob$ represents the output probabilities of various water types as predicted by the classifier, $R$ is the Renset34~\cite{7780459}, and $\theta$ represents the model weights of $R$ pre-trained on the SUIEBD dataset~\cite{LI2020107038}. This approach is inspired by the Haze-Line~\cite{berman2017diving} method, which classifies the underwater environment into nine distinct categories. The generation of water type prior knowledge is then computed using $Prob$ as follows:
\begin{equation}
    P_{water} = Expand(argmax\ Prob)
    \label{eq:p_water}
\end{equation}
where $P$ denotes the water type prior knowledge and $Expand$ represents the tensor expansion operation. To determine the most likely water type, the maximum probability value from $Prob$ is selected, identifying the category index with the highest probability. This index is then expanded to match the feature dimension of the FEN.
Building on \cite{wu2023learning, zhou2022conditional}, we utilize the intermediate layer features of the classifier to capture the fine-grained differences at a microscopic level. For a given underwater image $X$ and its corresponding classifier intermediate layer features $\{R_{i},\ i=0,1,2,3\}$, the generation of degradation prior knowledge is approached in two primary stages, starting with the fusion of information:
\begin{equation}
    P_{degrad}^{1} = Norm(\sum_{i=0}^{4}Conv(R_{i}^{2^{i}}))
\end{equation}
where $P_{degrad}^{1}$ denotes the first stage degradation prior knowledge. $Norm$ refers to Instance Normalization as defined in \cite{ulyanov2017instance}, and $Conv$ signifies a convolution operation with a kernel size of $1$. $R_{i}^{2^{i}}$ indicates that the $i$th feature of classifier is upsampled to $2^{i}$ times its original size. Following this initial stage, the process advances to feature refinement:
\begin{equation}
    P_{degrad} = ReLU(Conv(d_{degrade}^{1}))
    \label{eq:p_degrade}
\end{equation}

Unlike the frozen classifier, FEN performs feature enhancement based on task loss, which can capture the feature representation of the task demand well, such as brightness and color. Hence, the features of FEN are utilized as sample prior knowledge , the complete prior knowledge generation process can be formulated as:
\begin{equation}
    P_{sample} = F_{j}
    \label{eq:p_sample}
\end{equation}
\begin{equation}
    P = Norm(P_{water} + P_{degrad} + P_{sample})
    \label{eq:p}
\end{equation}
where $F_{j}$ represents the feature map of the $j$th layer of FEN. $P$ denotes the complete prior knowledge, which will be used as $Query$ in the Attention module below.

\begin{figure}[!t]
    \centering
    \includegraphics[width=0.7\linewidth, scale=0.4]{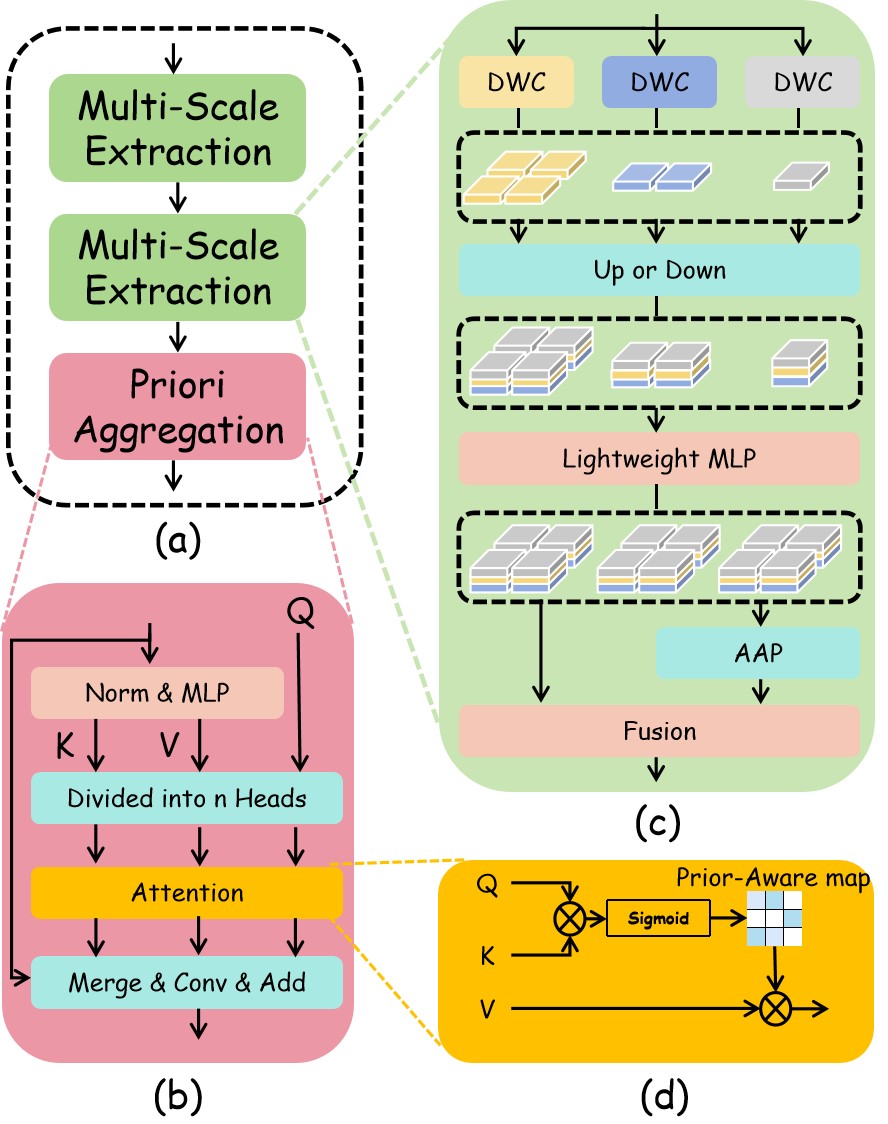}
    \caption{Structure of Multi-Scale Feature Aggregation (MSFA): Integrating Depth-Wise Convolution (DWC) and Adaptive Average Pooling (AAP).}
    \label{fig:msa}
\end{figure}

\begin{figure*}[t]
    \centering
    \begin{subfigure}[b]{1.0\textwidth}
        \centering
        \includegraphics[width=1.0\textwidth, height=0.24\textwidth]{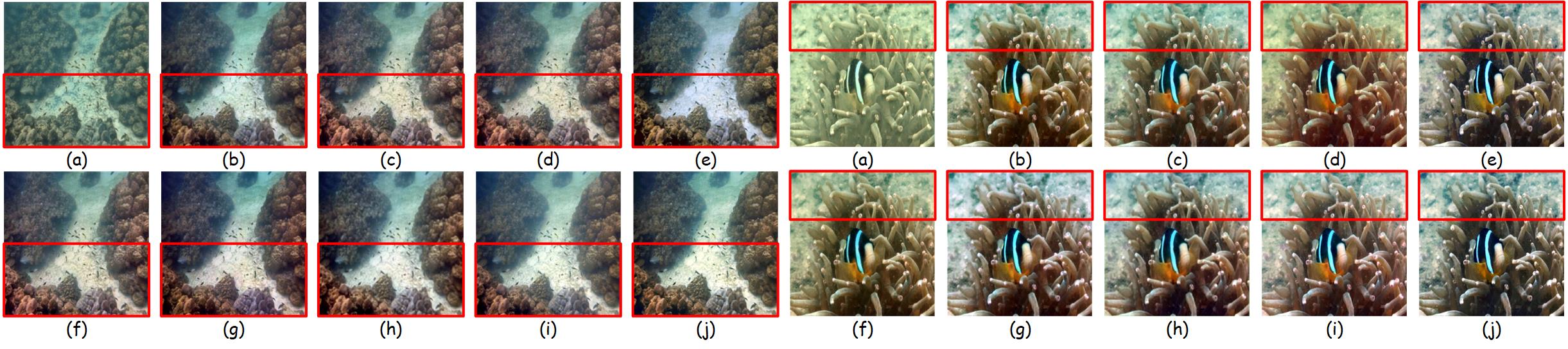}
        \caption{Visualization on UIEB}
    \end{subfigure}

    \begin{subfigure}[b]{1.0\textwidth}
        \centering
        \includegraphics[width=1.0\textwidth, height=0.24\textwidth]{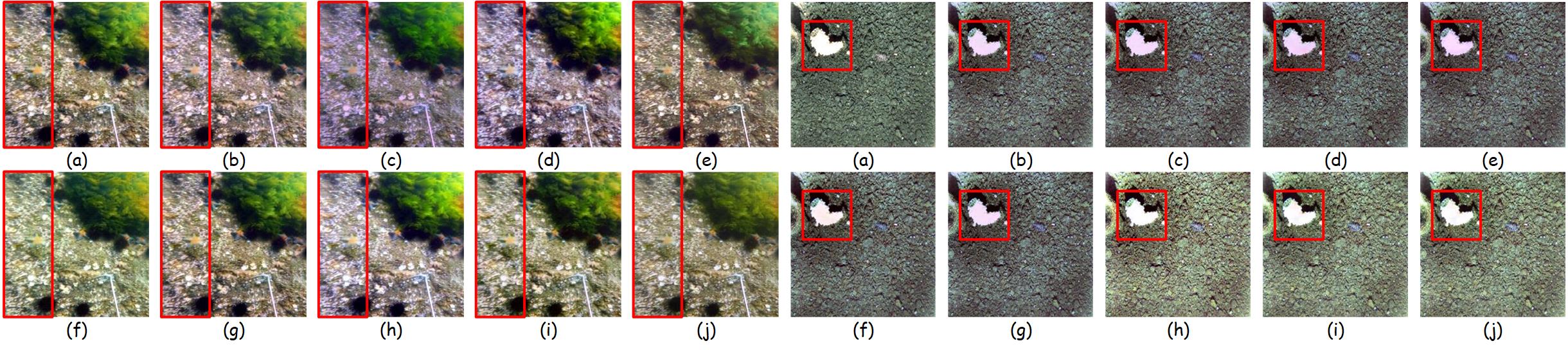}
        \caption{Visualization on LSUI}
    \end{subfigure}  
    \caption{Visualization of UIE methods. (a) Shallow-UWNet~\cite{naik2021shallow}, (b) NAFNet~\cite{chen2022simple} (c) Restormer~\cite{zamir2022restormer}, (d) FiveAPLUS~\cite{jiang2023five}, (e) FFANet~\cite{qin2020ffa}, (f) Shallow-UWNet + IA2U, (g) NAFNet + IA2U, (h) Restormer + IA2U, (i) FiveAPLUS + IA2U, (j) FFANet + IA2U.}
    \label{fig:uie}
\end{figure*}

\subsection{Multi-scale feature aggregation}
\cite{lin2023scale, hashmi2023featenhancer, cai2018cascade} has demonstrated the important role of multi-scale structural information in OD tasks. To enhance the performance of the IA2U plugin in OD, we developed a Multi-Scale Feature Aggregation module (MSFA) in \cref{fig:msa} (a). MSFA consists of two core modules: Multi-Scale Feature Extraction (MSFE) and Prior Aggregation (PA). The MSFE module is responsible for capturing and processing features across different scales, ensuring a comprehensive representation of the input data. The PA module focuses on combining the extracted multi-scale features with prior knowledge to improve the accuracy and robustness of OD.

{\bf Multi-Scale Feature Extraction:} 
As shown in \cref{fig:msa} (c), the MSFE module utilizes Depth-Wise Convolutions (DWC) with kernel sizes of $\{1, 3, 5\}$ to extract multi-scale features. The diversity of convolution kernels enables the model to effectively access feature information from different receptive fields and enhances its ability to model long-range dependencies. Subsequently, the extracted features at the $i$th scale, $S_{i}$, undergo feature alignment. Unlike traditional methods that choose a specific scale as the anchor point for alignment, we proposes a full-scale feature alignment strategy aimed at reducing information loss during resampling. When the receptive field of the scale feature $S_{i}$ is smaller than $S_{j}$ ($i < j$), upsampling is used; otherwise, downsampling is applied. The sampling function can be defined as:
\begin{equation}
    S_{i}^{j} = Sample(S_{i}, j) =
    \begin{cases}
        Bilinear(S_{i}, j)& \text{i \textgreater\ j}\\
        S_{i}& \text{i = j} \\
        AvgPool(S_{i}, j)& \text{i \textless\ j}
    \end{cases}
\end{equation}
where $S_{i}$ denotes the feature at scale $i$, $j$ represents the target scale, and $Sample(S_{i},\ j)$ means  the alignment of the $i$-scale feature to the $j$-scale. The $Bilinear(S_{i},\ j)$ represents the dimension of $S_{i}$ that is upsampled to $j$ using bilinear interpolation. Similarly, $AvgPool$ represents average pooling downsampling. In particular, $(i, j) \in \{(x, y)\ |\ x,\ y\in \{1, 3, 5\}\}$. The three sets of scale features obtained at this point contain un-lost information about themselves and spliced feature information from the remaining scales. The next stage is to employ a lightweight MLP network for processing, which performs feature refinement and maintains consistent feature map sizes, can be modeled as:
\begin{equation}
    \{\hat{S_{1}}, \hat{S_{2}}, \hat{S_{3}}\} = \{MLP(\{Concat(S_{1}^{m}, S_{3}^{m}, S_{5}^{m})\})\}
\end{equation}
where $m \in \{1, 3, 5\}$. Then, the weight map is generated using Adaptive Average Pooling (AAP) and tensor expansion operations, denoted by Eq.(\ref{eq:9}):
\begin{equation}
    \{W_{1}, W_{2}, W_{3}\} = Expand(AAP( \{\hat{S_{1}}, \hat{S_{2}}, \hat{S_{3}}\}))
    \label{eq:9}
\end{equation}
Finally, the output features are obtained by Eq.(\ref{eq:msa}), as follows:
\begin{equation}
    F_{out}^{1} = \sum_{i\ =\ 0}^{3} \hat{S_{i}} \times W_{i}
    \label{eq:msa}
\end{equation}

{\bf Priori Aggregation:} 
This study draws inspiration from Transformer-based methods \cite{vaswani2017attention} as discussed in \cite{dosovitskiy2020image, liu2021swin}, and employs a self-attention-like mechanism to utilize the prior knowledge $P$ from \cref{eq:p} for feature aggregation. Our innovation using the prior knowledge of water type, degradation, and samples as $Query$. As shown in \cref{fig:msa} (b), the feature $F_{out}^{1}$ obtained from \cref{eq:msa} is normalized and linearly transformed to generate $Key$ and $Value$. $Q,\ K,$ and $V$ are partitioned into $n$ segments for the attention computation, based on \cref{fig:msa} (d). Inspired by \cite{9930878, hashmi2023featenhancer, zamir2022restormer}, we did not adopt the traditional self-attention approach. Instead, we engage in an element-wise interaction between $P$ and $K$, and use the $sigmoid$ function to generate activation maps $M$. Subsequently, through the interaction between $M$ and $V$, output features are obtained under the condition of the prior $P$. This process can be described as the follows:
\begin{equation}
    M = Sigmoid(LT_{k}(F_{out}^{1}) \times P)
\end{equation}
\begin{equation}
    F_{out}^{2} = M \times LT_{v}(F_{out}^{1})
\end{equation}
where $LT_{k}$ and $LT_{v}$ represent the linear transformations corresponding to $K$ and $V$, respectively. Meanwhile, $P$ and $F_{out}^{1}$ are defined in \cref{eq:p} and \cref{eq:msa}, respectively.

\section{Experiment}
\label{sec:exp}

\subsection{Implementation Details}
For our experiments, we employed PyTorch version $1.12.1$ and the MMDetection framework~\cite{chen2019mmdetection} as our software platform. Training was conducted on four NVIDIA A800 GPUs, each with 80G of memory. To ensure consistency across experiments, we set the random seed to 0.

\begin{table}[htbp]
  \centering
  \begin{tabular}{@{}lcc@{}}
    \toprule
    Method & $AP(\%)$ & $AP_{50}(\%)$ \\
    \midrule
    Faster R-CNN & 49.1 & 80.3 \\
    Faster R-CNN + IA2U & 52.3 (\textcolor{red}{+ 3.2}) & 81.9 (\textcolor{red}{+ 1.6})\\
    Cascade R-CNN & 53.8 & 81.4\\
    Cascade R-CNN + IA2U & 54.7 (\textcolor{red}{+ 0.9}) & 81.6 (\textcolor{red}{+ 0.2})\\
    RatinaNet & 48.0 & 77.8\\
    RatinaNet + IA2U & 51.2 (\textcolor{red}{+ 3.2}) & 81.3 (\textcolor{red}{+ 3.5})\\
    ATSS & 53.9 & 82.2\\
    ATSS + IA2U & 54.6 (\textcolor{red}{+ 0.7}) & 82.8 (\textcolor{red}{+ 0.6}) \\
    TOOD & 55.3 & 83.1\\
    TOOD + IA2U & 56.1 (\textcolor{red}{+ 0.8}) & 83.1 \\
    PAA & 53.5 & 82.2\\
    PAA + IA2U & 55.6 (\textcolor{red}{+ 2.1}) & 83.3 (\textcolor{red}{+ 1.1})\\
    \bottomrule
  \end{tabular}
  \caption{Enhanced object detection performance using IA2U on the RUOD~\cite{fu2023rethinking}. Performance gains by IA2U indicated in \textcolor{red}{Red}. The in-air detectors were Faster R-CNN\cite{7485869}, Cascade R-CNN\cite{cai2018cascade}, RatinaNet\cite{8237586}, ATSS\cite{zhang2020bridging}, TOOD\cite{9710724}, and PAA\cite{kim2020probabilistic}.
  }
  \label{tab:det}
\end{table}

\begin{table*}[htbp]
\centering
\scalebox{0.8}{
\begin{tabular}{lccccc}
\toprule
Method               & PSNR$\uparrow$               & SSIM$\uparrow$              & LPIPS$\downarrow$             & NIQE$\downarrow$              & UCIQE$\uparrow$            \\
\midrule
Shallow              & 18.7630            & 0.8535            & 0.2196            & 4.8340            & 0.5481           \\
Shallow + IA2U      & 24.0022 (\textcolor{red}{+ 5.2392}) & 0.9135 (\textcolor{red}{+ 0.0600}) & 0.1038 (\textcolor{red}{- 0.1158}) & 4.6088 (\textcolor{red}{- 0.2252}) & 0.6031 (\textcolor{red}{+ 0.0550})\\
FFANet         & 23.6928            & 0.9189            & 0.1045            & 4.2994            & 0.6014           \\
FFANet + IA2U & 24.2774 (\textcolor{red}{+ 0.5846})   & 0.9189 & 0.1034 (\textcolor{red}{- 0.0011}) & 4.2566 (\textcolor{red}{- 0.0428}) & 0.6132 (\textcolor{red}{+ 0.0118})\\ 
NAFNet               & 23.0067            & 0.8900            & 0.1388            & 5.7542            & 0.5923           \\
NAFNet + IA2U       & 24.6099 (\textcolor{red}{+ 1.6032}) & 0.9262 (\textcolor{red}{+ 0.0362}) & 0.0894 (\textcolor{red}{- 0.0494}) & 4.5329 (\textcolor{red}{- 1.2213}) & 0.6020 (\textcolor{red}{+ 0.0097})\\
Restormer            & 23.9995            & 0.9201            & 0.0964            & 4.5216            & 0.5991           \\
Restormer + IA2U    & 25.5448 \textcolor{red}{(+ 1.5453}) & 0.9312 (\textcolor{red}{+ 0.0111}) & 0.0793 (\textcolor{red}{- 0.1710})  & 4.5832 (\textcolor{blue}{+ 0.0616}) & 0.6092 (\textcolor{red}{+ 0.0101})\\
FiveAPLUS         & 22.9076            & 0.8956            & 0.1358            & 4.3428            & 0.6106           \\
FiveAPLUS + IA2U & 24.7736 (\textcolor{red}{+ 1.8660})   & 0.9209 (\textcolor{red}{+ 0.0253}) & 0.0923 (\textcolor{red}{- 0.0435}) & 4.5755 (\textcolor{blue}{+ 0.2327}) & 0.6005 (\textcolor{blue}{- 0.0101})\\  
\bottomrule
\end{tabular}
}
\caption{Qualitative analysis on UIEBD~\cite{8917818}. Performance increase indicated in \textcolor{red}{Red}, and decrease in \textcolor{blue}{Blue}. FFANet\cite{qin2020ffa}, NAFNet\cite{chen2022simple}, and Restormer\cite{zamir2022restormer} are dehaze models; Shallow\cite{naik2021shallow} and FiveAPLUS\cite{jiang2023five} are UIE models without underwater priors.}
\label{tab:uieb}
\end{table*}

\begin{table*}[htbp]
\centering
\scalebox{0.8}{
\begin{tabular}{lccccc}
\toprule
Method               & PSNR$\uparrow$               & SSIM$\uparrow$              & LPIPS$\downarrow$             & NIQE$\downarrow$              & UCIQE$\uparrow$            \\
\midrule
Shallow              & 21.8756            & 0.8298            & 0.2249            & 4.7123            & 0.5578           \\
Shallow + IA2U      & 25.7063 (\textcolor{red}{+ 3.8307}) & 0.8693 (\textcolor{red}{+ 0.0395}) & 0.1474 (\textcolor{red}{- 0.0775}) & 4.6688 (\textcolor{red}{- 0.0435}) & 0.5926 (\textcolor{red}{+ 0.0348})\\
FFANet         & 27.4333            & 0.8848            & 0.1247            & 4.6740            & 0.5920          \\
FFANet + IA2U & 27.4616 (\textcolor{red}{+ 0.0283})   & 0.8850 (\textcolor{red}{+ 0.0002}) & 0.1243 (\textcolor{red}{- 0.0004}) & 4.6766 (\textcolor{blue}{+ 0.0026}) & 0.5928 (\textcolor{red}{+ 0.0020})\\ 
NAFNet               & 27.4155            & 0.8831            & 0.1287            & 4.7940            & 0.5948           \\
NAFNet + IA2U       & 27.6602 (\textcolor{red}{+ 0.2447}) & 0.8866 (\textcolor{red}{+ 0.0035}) & 0.1203 (\textcolor{red}{- 0.0084}) & 4.7489 (\textcolor{red}{- 0.0451}) & 0.5949 (\textcolor{red}{+ 0.0001})\\
Restormer            & 28.1776            & 0.8896            & 0.1142            & 4.6685            & 0.5957           \\
Restormer + IA2U    & 28.2096 \textcolor{red}{(+ 0.0320}) & 0.8908 (\textcolor{red}{+ 0.0012}) & 0.0793 (\textcolor{red}{- 0.0016})  & 4.6648 (\textcolor{blue}{+ 0.0037}) & 0.5968 (\textcolor{red}{+ 0.0011})\\
FiveAPLUS         & 23.0286            & 0.8443            & 0.2063            & 4.7906            & 0.5922           \\
FiveAPLUS + IA2U & 23.2864 (\textcolor{red}{+ 0.2578})   & 0.8511 (\textcolor{red}{+ 0.0068}) & 0.1880 (\textcolor{red}{- 0.0183}) & 4.8081 (\textcolor{blue}{+ 0.0175}) & 0.5908 (\textcolor{blue}{- 0.0014})\\       
\bottomrule
\end{tabular}
}
\caption{Qualitative analysis on LSUI~\cite{10129222}. Performance increase indicated 
 in \textcolor{red}{Red}, and decrease in \textcolor{blue}{Blue}. FFANet\cite{qin2020ffa}, NAFNet\cite{chen2022simple}, and Restormer\cite{zamir2022restormer} are dehaze models; Shallow\cite{naik2021shallow} and FiveAPLUS\cite{jiang2023five} are UIE models without underwater priors.}
\label{tab:lsui}
\end{table*}

{\bf Classifier Setup.} For our water-type classifier, we opted for ResNet34~\cite{7780459}, which was trained over 200 epochs on the SUIEBD dataset~\cite{LI2020107038}. We employed the AdamW optimizer for the classifier, setting the weight decay to $0.05$ and betas to $(0.9, 0.95)$. The learning rate was configured to increase linearly from $1 \times e^{-9}$ to $2.4 \times e^{-3}$ during the initial 20 epochs, followed by a gradual decline following a cosine decay strategy. Cross-entropy loss was utilized for constraint, leading to a final classifier that achieved the $Top1$ accuracy of $96.4\%$ on the test set.

{\bf Underwater Image Enhancement Setup.} The setup for UIE experiments followed a similar configuration to that of the classifier. However, there are distinctions in the learning rate parameters; the minimum learning rate was set at $1 \times e^{-6}$ and the maximum at $2 \times e^{-4}$. Additionally, to mitigate the impact of resolution changes on UIE performance, we implemented progressive training strategies. This involved adjusting the image size from $256 \times 256$ to $512 \times 512$ starting from the 150th epoch. Finally, the total loss function for UIE was defined as follows:
\begin{equation}
L_{total} = 1.0 \times L_{1} + 0.1 \times L_{ssim}
\end{equation}
where $L_1$ denotes the L1 loss and $L_{ssim}$ represents the structural similarity index measure loss.

{\bf Underwater Object Detection Setup.} 
For our UOD experiments, we utilized the MMDetection framework~\cite{chen2019mmdetection} as the basis for all object detection models. The experiments were conducted with an image size of $640 \times 640$ and a batch size of $64$. The optimizer chosen for this task was Stochastic Gradient Descent (SGD). The learning rate was configured to gradually increase from $1 \times e^{-3}$ to $2 \times e^{-2}$ over the first 500 iterations, providing a controlled ramp-up in the training process.

\subsection{Qualitative Analysis}
To evaluate the effectiveness of IA2U, we selected UIE and UOD as representative tasks for upstream and downstream processes, respectively. Extensive experiments were conducted to assess the performance of IA2U. This qualitative analysis aimed to provide a comprehensive understanding of how IA2U enhances the quality and accuracy of underwater imaging and detection tasks.

\begin{figure}[htbp]
    \centering
    \begin{subfigure}[b]{0.8\linewidth}
        \centering
        \includegraphics[width=1\linewidth]{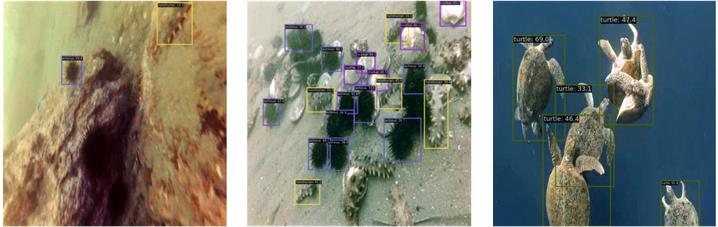}
        \caption{Semi-UIR + ATSS}
        \label{fig1:det_semi}
    \end{subfigure}
    \begin{subfigure}[b]{0.8\linewidth}
        \centering
        \includegraphics[width=1\linewidth]{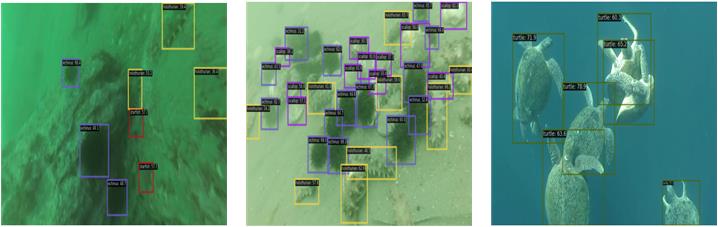}
        \caption{IA2U + ATSS}
        \label{fig1:det_ours}
    \end{subfigure}
    \caption{Comparison of object detection results. (a) UIE with Semi-UIR ~\cite{huang2023contrastive} vs. (b) IFE using IA2U, employing ATSS~\cite{zhang2020bridging} as the detector.
    }
    \label{fig:det}
\end{figure}

{\bf Underwater Object Detection Analysis.} For our UOD study, we selected a range of base methods including Faster R-CNN~\cite{7485869}, Cascade R-CNN~\cite{cai2018cascade}, RetinaNet~\cite{8237586}, ATSS~\cite{zhang2020bridging}, TOOD~\cite{9710724}, and PAA~\cite{kim2020probabilistic}, all trained on the latest RUOD dataset~\cite{fu2023rethinking}. Additionally, IA2U was implemented as a pre-module for these methods to further assess its effect. The comparative analysis from these two sets of experiments was conducted to evaluate the benefits of IA2U in the UOD task. According to the results presented in~\cref{tab:det}, it is evident that the in-air OD models, post integration with IA2U, show improved performance on the underwater dataset, with an average increase of $1.8\%$ in $AP$ (Average Precision @ IoU=$0.50:0.95$) and $1.2\%$ in $AP_{50}$ (Average Precision @ IoU=$0.50$).

Furthermore, we conducted a visual comparison between IA2U and other IE methods in the context of an OD task. We chose the state-of-the-art  method, Semi-UIR~\cite{huang2023contrastive}, for comparison, and ATSS~\cite{zhang2020bridging} as the detector. As illustrated in \cref{fig:det}, while Semi-UIR improved color drift and blurriness, it negatively affected the accuracy of UOD. This observation further corroborates the notion that IE does not necessarily translate to benefits in downstream OD tasks, as discussed in prior studies~\cite{9832540, hashmi2023featenhancer}.

\begin{table}[htbp]
  \centering
  \begin{tabular}{@{}cccccc@{}}
    \toprule
    W & D & S & FS & PSNR & SSIM \\
    \midrule
     \checkmark & \checkmark & \checkmark & \checkmark & \colorbox{red!50}{\textbf{24.002}} & \colorbox{red!50}{\textbf{0.914}} \\
     & \checkmark & \checkmark & \checkmark & 23.455 (\textcolor{blue}{- 0.547}) & 0.902 (\textcolor{blue}{- 0.012})\\
    \checkmark & & \checkmark & \checkmark & 23.253 (\textcolor{blue}{- 0.749}) & 0.908 (\textcolor{blue}{- 0.006})\\
    \checkmark & \checkmark & & \checkmark & 23.393 (\textcolor{blue}{- 0.609}) & 0.901 (\textcolor{blue}{- 0.013})\\
    \checkmark & \checkmark & \checkmark & & 23.314 (\textcolor{blue}{- 0.688}) & 0.901 (\textcolor{blue}{- 0.013})\\
    \bottomrule
  \end{tabular}
  \caption{Assessing the impact of Water (W), Degradation (D), Sample (S) prior knowledge, and Full-Scale (FS) feature alignment, with scale 3 as the anchor point in Non-FS cases.
  }
  \label{tab:ab}
\end{table}

{\bf Underwater Image Enhancement Analysis.} 
We selected several state-of-the-art models as baselines for UIE, including Shallow-UWNet~\cite{naik2021shallow}, NAFNet~\cite{chen2022simple}, Restormer~\cite{zamir2022restormer}, FiveAPLUS~\cite{jiang2023five}, and FFANet~\cite{qin2020ffa}. The evaluation was carried out on two datasets: the smaller UIEBD dataset~\cite{8917818} and the larger LSUI dataset~\cite{10129222}. For quantitative analysis, we used full reference metrics such as Peak Signal-to-Noise Ratio (PSNR), Structural Similarity (SSIM), and Learned Perceptual Image Patch Similarity (LPIPS)~\cite{zhang2018unreasonable}, and no reference metrics Natural Image Quality Evaluator (NIQE)~\cite{mittal2012making}, Underwater Color Image Quality Evaluation (UCIQE)~\cite{yang2015underwater} are selected for qualitative analysis. 

We randomly divided 90 images to create a test set, following to the UIEB dataset. The outcomes of these tests are presented in \cref{tab:uieb}. It is clear that, integrating IA2U into in-air models significantly enhances UIE performance. It's worth noting that the NAFNet, Restormer, and FiveAPLUS methods have effectively improved image quality and outperformed Semi-UIR \cite{huang2023contrastive} ($PSNR=24.59$, $SSIM=0.901$). Additionally, IA2U yielded significant results on the LSUI dataset, and the detailed scores in \cref{tab:lsui}.

\cref{fig:uie} illustrates the visual effects, it is evident that due to the lack of underwater environmental information, the in-air model struggles to correct the color shift in underwater images and may even introduce unnatural colors. IA2U serving as an UIE module, enables the in-air model to adapt to the underwater environment, improved quality of underwater images.

\subsection{Ablation Study}

We conducted ablation experiments using the Shallow-UWNet~\cite{naik2021shallow} on the UIEB dataset. The main ablation targets were the sample selection method and the multi-scale feature alignment method.

\begin{figure}[htbp]
    \centering
    \includegraphics[width=0.8\linewidth]{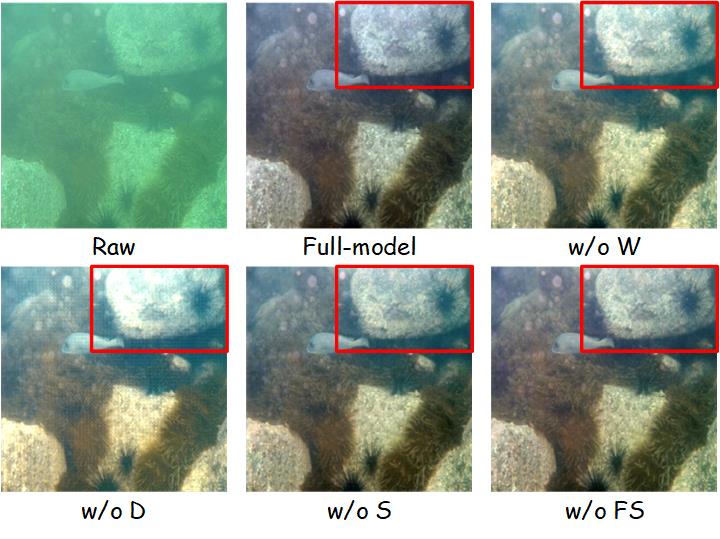}
    \caption{Visualization of ablation experiments. $w/o$ stands for {\bf WITHOUT}. $W$, $D$, and $S$ denote water type, degradation, and sample prior, respectively. $FS$ stands for full-scale feature alignment.}
    \label{fig:ab}
\end{figure}

The results of the ablation experiments are shown in \cref{tab:ab}, and all three prior knowledge selected by IA2U are valid, with the degradation prior having the most significant impact on the model. The full-scale alignment of the MSFA module is also better than the previous single-anchor alignment. The effect of prior knowledge on visualization is illustrated in \cref{fig:ab}.

\section{Conclusion}
\label{sec:conc}

In this paper, we propose IA2U, which is a plug-and-play feature enhancement plugin specifically designed for underwater environments, aimed at adapting in-air models for underwater environment. IA2U enhances the model's generalization capabilities by integrating underwater environmental information. Utilizing a full-scale feature fusion strategy, it minimizes the loss during multi-scale feature fusion, effectively extracting task-level features. IA2U demonstrates superior performance in UIE and UOD tasks.

{
    \small
    \bibliographystyle{ieeenat_fullname}
    \bibliography{arxiv}

\begin{thebibliography}{44}
\providecommand{\natexlab}[1]{#1}
\providecommand{\url}[1]{\texttt{#1}}
\expandafter\ifx\csname urlstyle\endcsname\relax
  \providecommand{\doi}[1]{doi: #1}\else
  \providecommand{\doi}{doi: \begingroup \urlstyle{rm}\Url}\fi

\bibitem[Akkaynak and Treibitz(2019)]{akkaynak2019sea}
Derya Akkaynak and Tali Treibitz.
\newblock Sea-thru: A method for removing water from underwater images.
\newblock In \emph{CVPR}, pages 1682--1691, 2019.

\bibitem[Berman et~al.(2017)Berman, Treibitz, and Avidan]{berman2017diving}
Dana Berman, Tali Treibitz, and Shai Avidan.
\newblock Diving into haze-lines: Color restoration of underwater images.
\newblock In \emph{BMVC}, page~2, 2017.

\bibitem[Cai and Vasconcelos(2018)]{cai2018cascade}
Zhaowei Cai and Nuno Vasconcelos.
\newblock Cascade r-cnn: Delving into high quality object detection.
\newblock In \emph{CVPR}, pages 6154--6162, 2018.

\bibitem[Chen et~al.(2019)Chen, Wang, Pang, Cao, Xiong, Li, Sun, Feng, Liu, Xu, et~al.]{chen2019mmdetection}
Kai Chen, Jiaqi Wang, Jiangmiao Pang, Yuhang Cao, Yu Xiong, Xiaoxiao Li, Shuyang Sun, Wansen Feng, Ziwei Liu, Jiarui Xu, et~al.
\newblock Mmdetection: Open mmlab detection toolbox and benchmark.
\newblock \emph{arXiv preprint arXiv:1906.07155}, 2019.

\bibitem[Chen et~al.(2022)Chen, Chu, Zhang, and Sun]{chen2022simple}
Liangyu Chen, Xiaojie Chu, Xiangyu Zhang, and Jian Sun.
\newblock Simple baselines for image restoration.
\newblock In \emph{ECCV}, pages 17--33. Springer, 2022.

\bibitem[Dosovitskiy et~al.(2020)Dosovitskiy, Beyer, Kolesnikov, Weissenborn, Zhai, Unterthiner, Dehghani, Minderer, Heigold, Gelly, et~al.]{dosovitskiy2020image}
Alexey Dosovitskiy, Lucas Beyer, Alexander Kolesnikov, Dirk Weissenborn, Xiaohua Zhai, Thomas Unterthiner, Mostafa Dehghani, Matthias Minderer, Georg Heigold, Sylvain Gelly, et~al.
\newblock An image is worth 16x16 words: Transformers for image recognition at scale.
\newblock \emph{arXiv preprint arXiv:2010.11929}, 2020.

\bibitem[Feng et~al.(2021)Feng, Zhong, Gao, Scott, and Huang]{9710724}
Chengjian Feng, Yujie Zhong, Yu Gao, Matthew~R. Scott, and Weilin Huang.
\newblock Tood: Task-aligned one-stage object detection.
\newblock In \emph{ICCV}, pages 3490--3499, 2021.

\bibitem[Fu et~al.(2023)Fu, Liu, Fan, Chen, Fu, Yuan, Zhu, and Luo]{fu2023rethinking}
Chenping Fu, Risheng Liu, Xin Fan, Puyang Chen, Hao Fu, Wanqi Yuan, Ming Zhu, and Zhongxuan Luo.
\newblock Rethinking general underwater object detection: Datasets, challenges, and solutions.
\newblock \emph{Neurocomputing}, 517:\penalty0 243--256, 2023.

\bibitem[Guo et~al.(2023)Guo, Wu, Jin, Han, Chai, Zhang, and Li]{URanker}
Chunle Guo, Ruiqi Wu, Xin Jin, Linghao Han, Zhi Chai, Weidong Zhang, and Chongyi Li.
\newblock Underwater ranker: Learn which is better and how to be better.
\newblock In \emph{AAAI}, pages 720--709, 2023.

\bibitem[Hashmi et~al.(2023)Hashmi, Kallempudi, Stricker, and Afzal]{hashmi2023featenhancer}
Khurram~Azeem Hashmi, Goutham Kallempudi, Didier Stricker, and Muhammad~Zeshan Afzal.
\newblock Featenhancer: Enhancing hierarchical features for object detection and beyond under low-light vision.
\newblock In \emph{ICCV}, pages 6725--6735, 2023.

\bibitem[He et~al.(2016)He, Zhang, Ren, and Sun]{7780459}
Kaiming He, Xiangyu Zhang, Shaoqing Ren, and Jian Sun.
\newblock Deep residual learning for image recognition.
\newblock In \emph{CVPR}, pages 770--778, 2016.

\bibitem[Huang et~al.(2023)Huang, Wang, Liu, Chen, and Li]{huang2023contrastive}
Shirui Huang, Keyan Wang, Huan Liu, Jun Chen, and Yunsong Li.
\newblock Contrastive semi-supervised learning for underwater image restoration via reliable bank.
\newblock In \emph{CVPR}, pages 18145--18155, 2023.

\bibitem[Jaffe(1990)]{50695}
J.S. Jaffe.
\newblock Computer modeling and the design of optimal underwater imaging systems.
\newblock \emph{IEEE JOE}, 15\penalty0 (2):\penalty0 101--111, 1990.

\bibitem[Jiang et~al.(2023)Jiang, Ye, Bai, Chen, Chai, Jun, Liu, and Chen]{jiang2023five}
Jingxia Jiang, Tian Ye, Jinbin Bai, Sixiang Chen, Wenhao Chai, Shi Jun, Yun Liu, and Erkang Chen.
\newblock Five a+ network: You only need 9k parameters for underwater image enhancement.
\newblock \emph{arXiv preprint arXiv:2305.08824}, 2023.

\bibitem[Kang et~al.(2023)Kang, Jiang, Li, Ren, Liu, and Wang]{9895452}
Yaozu Kang, Qiuping Jiang, Chongyi Li, Wenqi Ren, Hantao Liu, and Pengjun Wang.
\newblock A perception-aware decomposition and fusion framework for underwater image enhancement.
\newblock \emph{IEEE TCSVT}, 33\penalty0 (3):\penalty0 988--1002, 2023.

\bibitem[Kim and Lee(2020)]{kim2020probabilistic}
Kang Kim and Hee~Seok Lee.
\newblock Probabilistic anchor assignment with iou prediction for object detection.
\newblock In \emph{ECCV}, pages 355--371. Springer, 2020.

\bibitem[Li et~al.(2020{\natexlab{a}})Li, Anwar, and Porikli]{LI2020107038}
Chongyi Li, Saeed Anwar, and Fatih Porikli.
\newblock Underwater scene prior inspired deep underwater image and video enhancement.
\newblock \emph{PR}, 98:\penalty0 107038, 2020{\natexlab{a}}.

\bibitem[Li et~al.(2020{\natexlab{b}})Li, Guo, Ren, Cong, Hou, Kwong, and Tao]{8917818}
Chongyi Li, Chunle Guo, Wenqi Ren, Runmin Cong, Junhui Hou, Sam Kwong, and Dacheng Tao.
\newblock An underwater image enhancement benchmark dataset and beyond.
\newblock \emph{IEEE TIP}, 29:\penalty0 4376--4389, 2020{\natexlab{b}}.

\bibitem[Li et~al.(2021)Li, Anwar, Hou, Cong, Guo, and Ren]{9426457}
Chongyi Li, Saeed Anwar, Junhui Hou, Runmin Cong, Chunle Guo, and Wenqi Ren.
\newblock Underwater image enhancement via medium transmission-guided multi-color space embedding.
\newblock \emph{IEEE TIP}, 30:\penalty0 4985--5000, 2021.

\bibitem[Lin et~al.(2017)Lin, Goyal, Girshick, He, and Dollár]{8237586}
Tsung-Yi Lin, Priya Goyal, Ross Girshick, Kaiming He, and Piotr Dollár.
\newblock Focal loss for dense object detection.
\newblock In \emph{ICCV}, pages 2999--3007, 2017.

\bibitem[Lin et~al.(2023)Lin, Wu, Chen, Huang, and Jin]{lin2023scale}
Weifeng Lin, Ziheng Wu, Jiayu Chen, Jun Huang, and Lianwen Jin.
\newblock Scale-aware modulation meet transformer.
\newblock In \emph{ICCV}, pages 6015--6026, 2023.

\bibitem[Liu et~al.(2022)Liu, Jiang, Yang, and Fan]{9832540}
Risheng Liu, Zhiying Jiang, Shuzhou Yang, and Xin Fan.
\newblock Twin adversarial contrastive learning for underwater image enhancement and beyond.
\newblock \emph{IEEE TIP}, 31:\penalty0 4922--4936, 2022.

\bibitem[Liu et~al.(2021)Liu, Lin, Cao, Hu, Wei, Zhang, Lin, and Guo]{liu2021swin}
Ze Liu, Yutong Lin, Yue Cao, Han Hu, Yixuan Wei, Zheng Zhang, Stephen Lin, and Baining Guo.
\newblock Swin transformer: Hierarchical vision transformer using shifted windows.
\newblock In \emph{ICCV}, pages 10012--10022, 2021.

\bibitem[Mittal et~al.(2012)Mittal, Soundararajan, and Bovik]{mittal2012making}
Anish Mittal, Rajiv Soundararajan, and Alan~C Bovik.
\newblock Making a “completely blind” image quality analyzer.
\newblock \emph{IEEE Sign. Process. Letters}, 20\penalty0 (3):\penalty0 209--212, 2012.

\bibitem[Naik et~al.(2021)Naik, Swarnakar, and Mittal]{naik2021shallow}
Ankita Naik, Apurva Swarnakar, and Kartik Mittal.
\newblock Shallow-uwnet: Compressed model for underwater image enhancement.
\newblock In \emph{AAAI}, pages 15853--15854, 2021.

\bibitem[Peng et~al.(2023)Peng, Zhu, and Bian]{10129222}
Lintao Peng, Chunli Zhu, and Liheng Bian.
\newblock U-shape transformer for underwater image enhancement.
\newblock \emph{IEEE TIP}, 32:\penalty0 3066--3079, 2023.

\bibitem[Qi et~al.(2022)Qi, Li, Zheng, Gao, Hou, and Sun]{9930878}
Qi Qi, Kunqian Li, Haiyong Zheng, Xiang Gao, Guojia Hou, and Kun Sun.
\newblock Sguie-net: Semantic attention guided underwater image enhancement with multi-scale perception.
\newblock \emph{IEEE TIP}, 31:\penalty0 6816--6830, 2022.

\bibitem[Qin et~al.(2020)Qin, Wang, Bai, Xie, and Jia]{qin2020ffa}
Xu Qin, Zhilin Wang, Yuanchao Bai, Xiaodong Xie, and Huizhu Jia.
\newblock Ffa-net: Feature fusion attention network for single image dehazing.
\newblock In \emph{AAAI}, pages 11908--11915, 2020.

\bibitem[Ren et~al.(2017)Ren, He, Girshick, and Sun]{7485869}
Shaoqing Ren, Kaiming He, Ross Girshick, and Jian Sun.
\newblock Faster r-cnn: Towards real-time object detection with region proposal networks.
\newblock \emph{IEEE TPAMI}, 39\penalty0 (6):\penalty0 1137--1149, 2017.

\bibitem[Tian et~al.(2019)Tian, Shen, Chen, and He]{9010746}
Zhi Tian, Chunhua Shen, Hao Chen, and Tong He.
\newblock Fcos: Fully convolutional one-stage object detection.
\newblock In \emph{ICCV}, pages 9626--9635, 2019.

\bibitem[Ulyanov et~al.(2017)Ulyanov, Vedaldi, and Lempitsky]{ulyanov2017instance}
Dmitry Ulyanov, Andrea Vedaldi, and Victor Lempitsky.
\newblock Instance normalization: The missing ingredient for fast stylization, 2017.

\bibitem[Vaswani et~al.(2017)Vaswani, Shazeer, Parmar, Uszkoreit, Jones, Gomez, Kaiser, and Polosukhin]{vaswani2017attention}
Ashish Vaswani, Noam Shazeer, Niki Parmar, Jakob Uszkoreit, Llion Jones, Aidan~N Gomez, {\L}ukasz Kaiser, and Illia Polosukhin.
\newblock Attention is all you need.
\newblock \emph{NeurIPS}, 30, 2017.

\bibitem[Wu et~al.(2023)Wu, Pan, Wang, Yang, Wei, Li, and Shen]{wu2023learning}
Yuhui Wu, Chen Pan, Guoqing Wang, Yang Yang, Jiwei Wei, Chongyi Li, and Heng~Tao Shen.
\newblock Learning semantic-aware knowledge guidance for low-light image enhancement.
\newblock In \emph{CVPR}, pages 1662--1671, 2023.

\bibitem[Yang and Sowmya(2015)]{yang2015underwater}
Miao Yang and Arcot Sowmya.
\newblock An underwater color image quality evaluation metric.
\newblock \emph{IEEE TIP}, 24\penalty0 (12):\penalty0 6062--6071, 2015.

\bibitem[Yuan et~al.(2021)Yuan, Cao, Cai, and Su]{9257110}
Jieyu Yuan, Wei Cao, Zhanchuan Cai, and Binghua Su.
\newblock An underwater image vision enhancement algorithm based on contour bougie morphology.
\newblock \emph{IEEE TGRS}, 59\penalty0 (10):\penalty0 8117--8128, 2021.

\bibitem[Yuan et~al.(2022)Yuan, Cai, and Cao]{TEBCF}
Jieyu Yuan, Zhanchuan Cai, and Wei Cao.
\newblock Tebcf: real-world underwater image texture enhancement model based on blurriness and color fusion.
\newblock \emph{IEEE TGRS}, 60:\penalty0 1--15, 2022.

\bibitem[Zamir et~al.(2022)Zamir, Arora, Khan, Hayat, Khan, and Yang]{zamir2022restormer}
Syed~Waqas Zamir, Aditya Arora, Salman Khan, Munawar Hayat, Fahad~Shahbaz Khan, and Ming-Hsuan Yang.
\newblock Restormer: Efficient transformer for high-resolution image restoration.
\newblock In \emph{CVPR}, pages 5728--5739, 2022.

\bibitem[Zhang et~al.(2018)Zhang, Isola, Efros, Shechtman, and Wang]{zhang2018unreasonable}
Richard Zhang, Phillip Isola, Alexei~A Efros, Eli Shechtman, and Oliver Wang.
\newblock The unreasonable effectiveness of deep features as a perceptual metric.
\newblock In \emph{CVPR}, pages 586--595, 2018.

\bibitem[Zhang et~al.(2020)Zhang, Chi, Yao, Lei, and Li]{zhang2020bridging}
Shifeng Zhang, Cheng Chi, Yongqiang Yao, Zhen Lei, and Stan~Z Li.
\newblock Bridging the gap between anchor-based and anchor-free detection via adaptive training sample selection.
\newblock In \emph{CVPR}, pages 9759--9768, 2020.

\bibitem[Zhang et~al.(2022)Zhang, Zhuang, Sun, Li, Kwong, and Li]{9788535}
Weidong Zhang, Peixian Zhuang, Hai-Han Sun, Guohou Li, Sam Kwong, and Chongyi Li.
\newblock Underwater image enhancement via minimal color loss and locally adaptive contrast enhancement.
\newblock \emph{IEEE TIP}, 31:\penalty0 3997--4010, 2022.

\bibitem[Zhou et~al.(2023{\natexlab{a}})Zhou, Li, Zhang, Yuan, Zhang, Cai, and Shi]{10177702}
Jingchun Zhou, Boshen Li, Dehuan Zhang, Jieyu Yuan, Weishi Zhang, Zhanchuan Cai, and Jinyu Shi.
\newblock Ugif-net: An efficient fully guided information flow network for underwater image enhancement.
\newblock \emph{IEEE TGRS}, 61:\penalty0 1--17, 2023{\natexlab{a}}.

\bibitem[Zhou et~al.(2023{\natexlab{b}})Zhou, Liu, Jiang, Ren, Lam, and Zhang]{zhou2023underwater}
Jingchun Zhou, Qian Liu, Qiuping Jiang, Wenqi Ren, Kin-Man Lam, and Weishi Zhang.
\newblock Underwater camera: Improving visual perception via adaptive dark pixel prior and color correction.
\newblock \emph{IJCV}, pages 1--19, 2023{\natexlab{b}}.

\bibitem[Zhou et~al.(2022)Zhou, Yang, Loy, and Liu]{zhou2022conditional}
Kaiyang Zhou, Jingkang Yang, Chen~Change Loy, and Ziwei Liu.
\newblock Conditional prompt learning for vision-language models.
\newblock In \emph{CVPR}, pages 16816--16825, 2022.

\bibitem[Zhuang et~al.(2022)Zhuang, Wu, Porikli, and Li]{9854113}
Peixian Zhuang, Jiamin Wu, Fatih Porikli, and Chongyi Li.
\newblock Underwater image enhancement with hyper-laplacian reflectance priors.
\newblock \emph{IEEE TIP}, 31:\penalty0 5442--5455, 2022.

\end{thebibliography}
}


\end{document}